\documentclass{article} %
\usepackage{iclr2021_conference,times}

\usepackage{amsmath,amsfonts,bm}

\def\eqref#1{equation~\ref{#1}}

\def\1{\bm{1}}

\DeclareMathAlphabet{\mathsfit}{\encodingdefault}{\sfdefault}{m}{sl}
\SetMathAlphabet{\mathsfit}{bold}{\encodingdefault}{\sfdefault}{bx}{n}

\usepackage{hyperref}

\usepackage{booktabs}
\usepackage{cleveref}
\usepackage{floatrow}
\usepackage{graphicx}
\usepackage{siunitx}
\usepackage{tabularx}
\usepackage{url}
\usepackage{xcolor}
\usepackage{xspace}
\usepackage{wrapfig}

\usepackage{tikz,graphics,color,float,epsf,caption,subcaption}

\title{\textsc{SI-Score}: An image dataset for fine-grained analysis of robustness to object location, \\ rotation and size}

\author{Jessica Yung,\hspace{0.03cm} Rob Romijnders,\hspace{0.03cm} Alexander Kolesnikov, \hspace{0.03cm} Lucas Beyer,\hspace{0.03cm} Josip Djolonga, \\
\textbf{Neil Houlsby,\hspace{0.03cm} Sylvain Gelly,\hspace{0.03cm} Mario Lucic,\hspace{0.03cm} Xiaohua Zhai} \thanks{Please send correspondence to \texttt{j.yung357@gmail.com}.} \\
Google Research, Brain Team\\
}

\newcommand{\imagenet}{\textsc{ImageNet}\xspace}

\iclrfinalcopy %
\begin{document}

\maketitle

\begin{abstract}
Before deploying machine learning models it is critical to assess their robustness. 
In the context of deep neural networks for image understanding, changing the object location, rotation and size may affect the predictions in non-trivial ways. In this work we perform a fine-grained analysis of robustness with respect to these factors of variation using \textsc{SI-Score}, a synthetic dataset. In particular, we investigate ResNets, Vision Transformers and CLIP, and identify interesting qualitative differences between these.
\end{abstract}

\section{Introduction}\label{sec:introduction}

In practice we would like to deploy models which are robust to certain changes in their input. For some of these factors, such as weather conditions, compression artifacts, or even different object orientations, existing datasets can be readily applied to quantify models’ robustness, e.g.\ \citet{barbu2019objectnet, hendrycks2018benchmarking}. However, for other important factors such as object size or location, the effect on model performance had not yet been quantified prior to our work. This is particularly concerning because many popular image datasets suffer from \textit{photographer's bias}~\citep{DBLP:conf/cvpr/TorralbaE11}, where objects appear mostly in the center of the image. %

\begin{figure*}
    \centering
    \begin{subfigure}[b]{0.49\textwidth}
        \centering
        \includegraphics[width=0.3\linewidth]{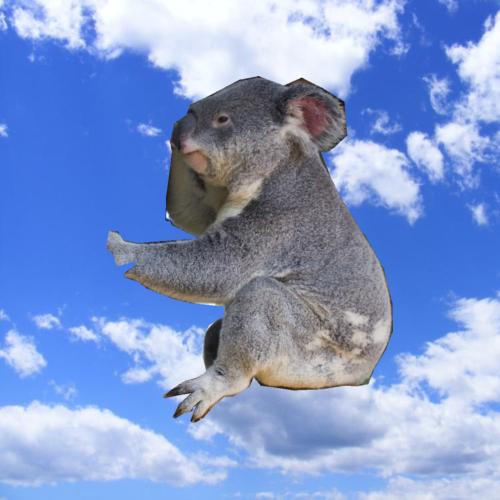}
        \includegraphics[width=0.3\linewidth]{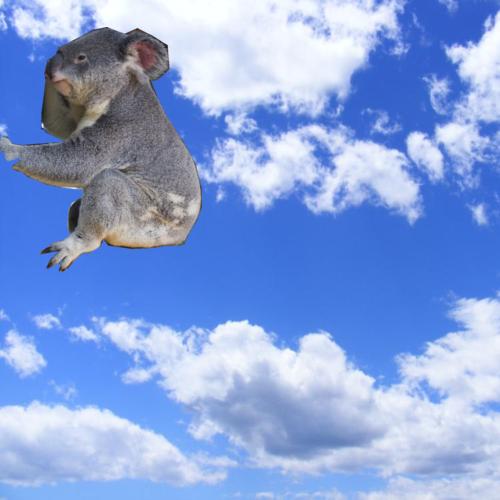}
        \includegraphics[width=0.3\linewidth]{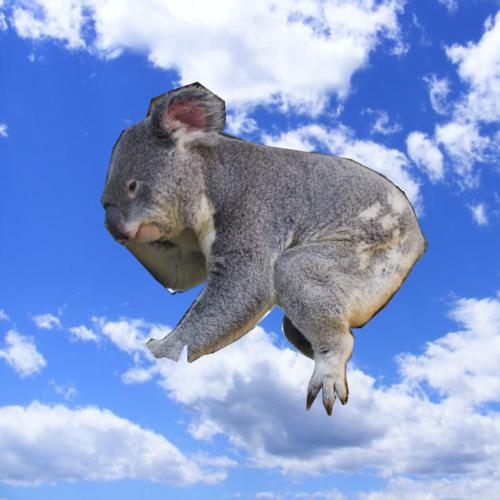}\\
        \includegraphics[width=0.3\linewidth]{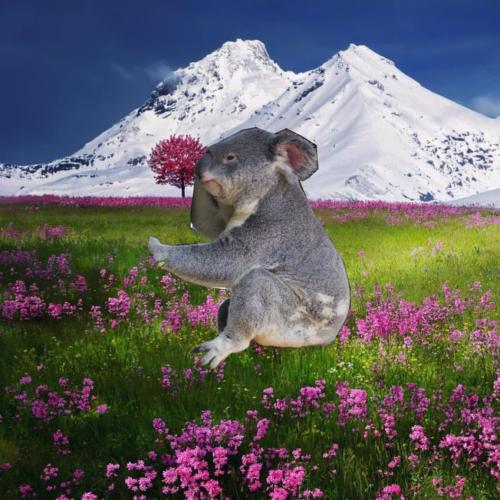}
        \includegraphics[width=0.3\linewidth]{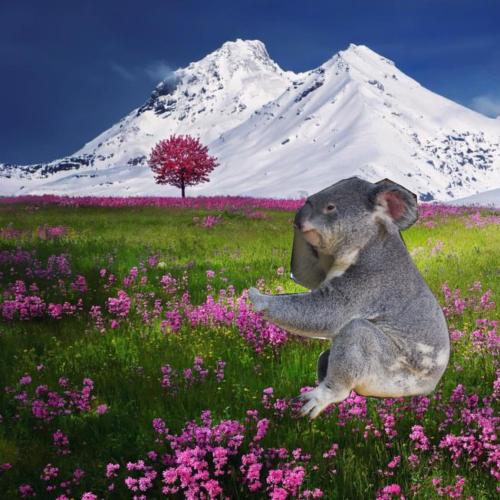}
        \includegraphics[width=0.3\linewidth]{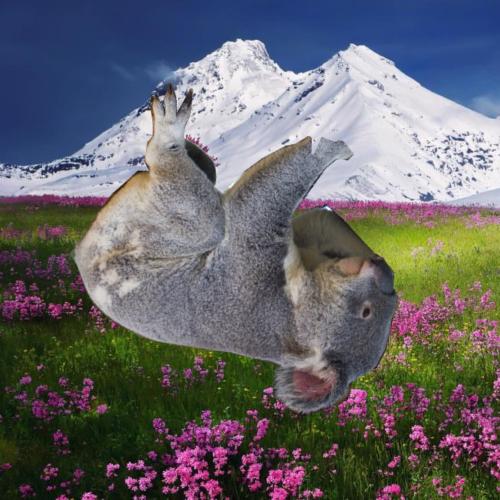}  
    \end{subfigure}\hfill
    \begin{subfigure}[b]{0.49\textwidth}  
        \centering 
        \includegraphics[width=0.2\linewidth]{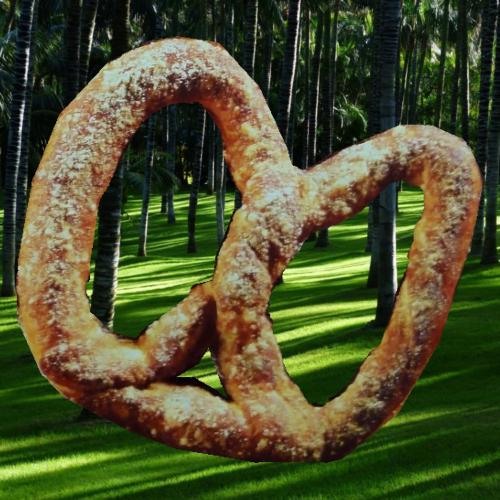}
        \includegraphics[width=0.2\linewidth]{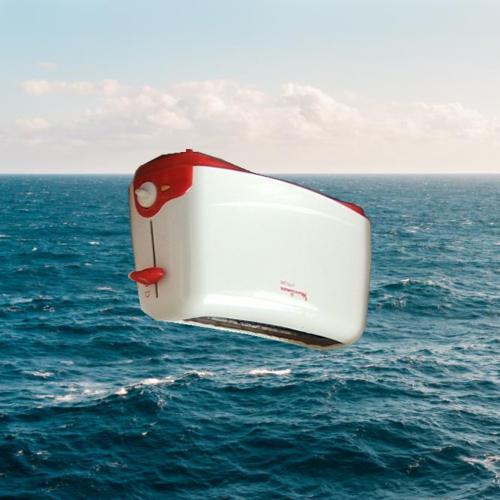}
        \includegraphics[width=0.2\linewidth]{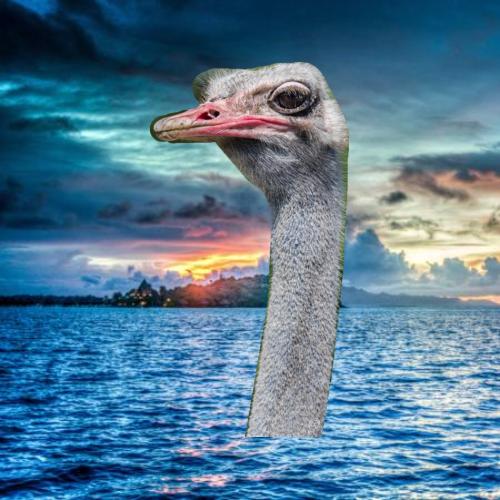}
        \includegraphics[width=0.2\linewidth]{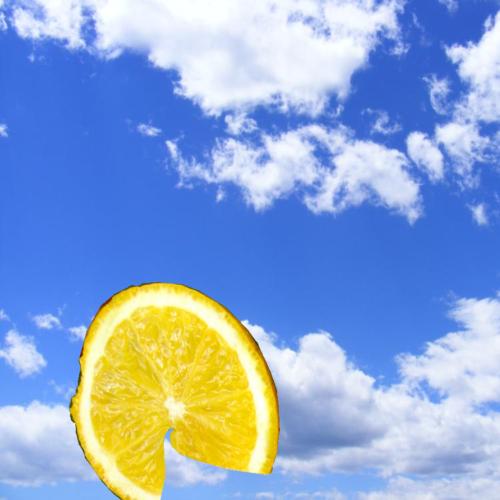} \\
        \includegraphics[width=0.2\linewidth]{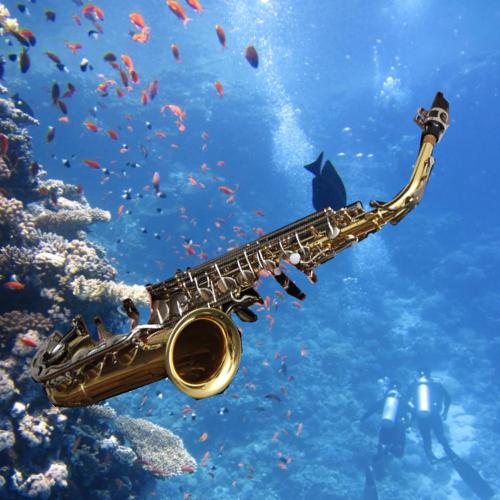}
        \includegraphics[width=0.2\linewidth]{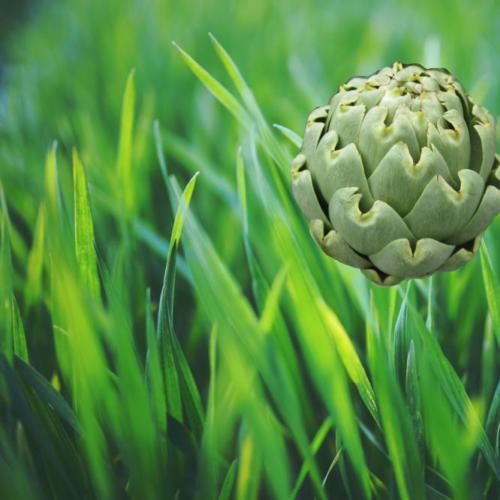}
        \includegraphics[width=0.2\linewidth]{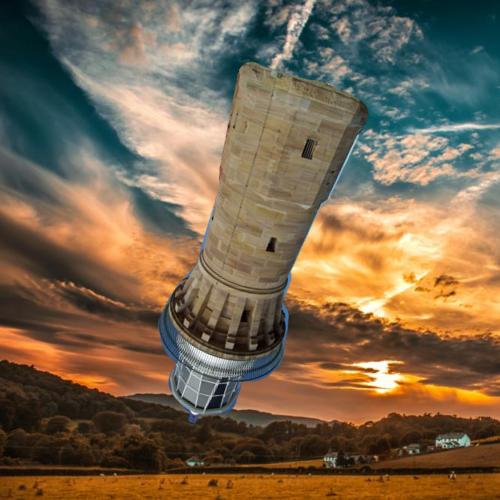}
        \includegraphics[width=0.2\linewidth]{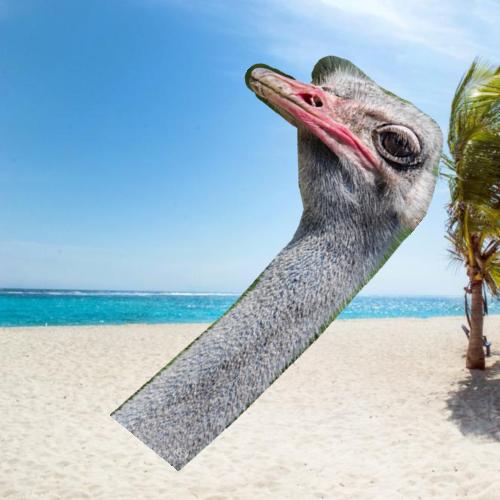} \\
        \includegraphics[width=0.2\linewidth]{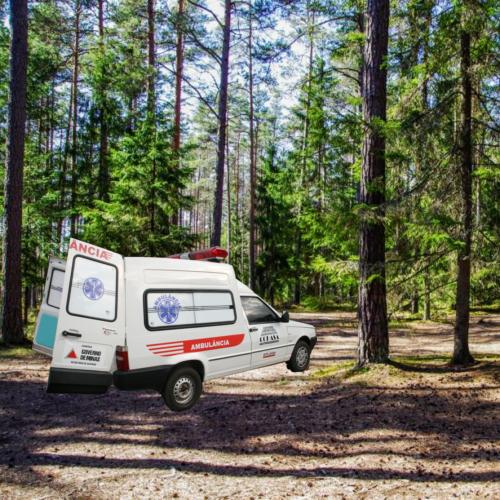}
        \includegraphics[width=0.2\linewidth]{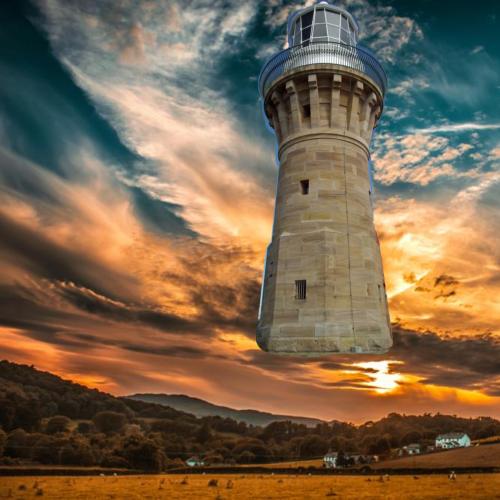}
        \includegraphics[width=0.2\linewidth]{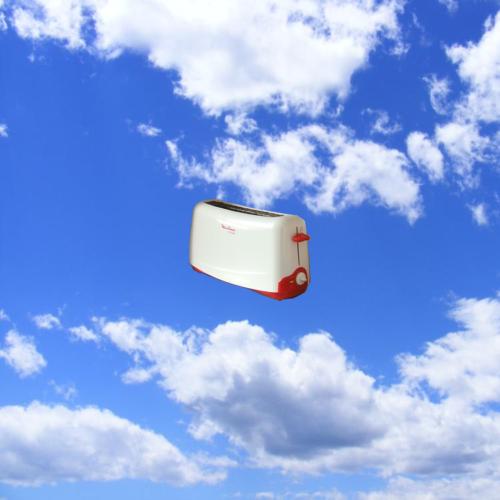}
        \includegraphics[width=0.2\linewidth]{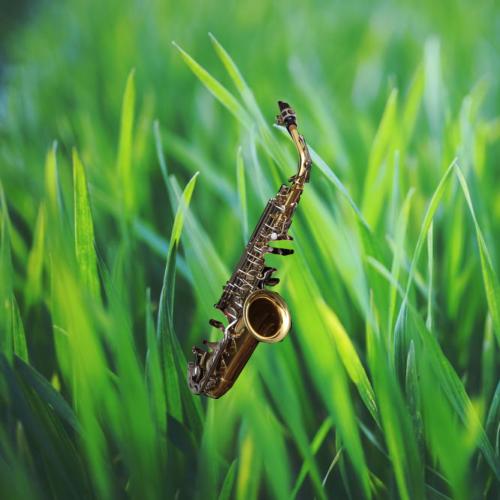}
    \end{subfigure}
    \vspace{1.5mm}
    \caption{Sample images from our synthetic dataset. \textbf{Left}: We paste the same foreground-background combination with the object in different sizes, locations and rotation angles. \textbf{Right}: We consider 614 foreground objects from 62 classes and 867 backgrounds, and vary the object location, rotation angle, and object size for a total of 
    \num{611608} images.} \label{fig:synth_examples} \vspace{-4mm}
\end{figure*}

In previous work \citep{djolonga2020robustness}, we open-sourced a synthetic dataset for fine-grained evaluation: \textsc{SI-Score} (Synthetic Interventions on Scenes for Robustness Evaluation). In a nutshell, we paste a large collection of objects onto uncluttered backgrounds (\Cref{fig:synth_examples}), and can thus conduct controlled studies by systematically varying the object class, size, location, and orientation. We also provided extendable code for researchers to generate similar synthetic datasets and analyse the results.\footnote{The synthetic dataset and code used to generate the dataset are available on GitHub and CVF at  \href{https://github.com/google-research/si-score}{\texttt{https://github.com/google-research/si-score}}.}

In this work, we take a step forward and identify interesting qualitative differences between model classes. In particular, we investigate how models based on convolutions (ResNets \citep{he2016deep}) compare to models based on attention, specifically Vision Transformers (ViT) \citep{dosovitskiy2020vit}. Moreover, we evaluate CLIP \citep{2021clip}, a model trained jointly on text and images on large-scale web data and evaluated zero-shot. %

\noindent\textbf{Related work}\quad Creating synthetic datasets by pasting objects onto backgrounds has been used for training \citep{zhao2020distilling, dwibedicutpaste2017, ghiasi2020copypaste} and evaluating models \citep{bit}, but previous works do not systematically vary object size, location or orientation, or analyse translation and rotation robustness only at the image level~\citep{engstrom2017}. GANs have also been used to generate counterfactual images to detect bias, specifically to evaluate the effects of features such as makeup or beards on classifiers \citep{denton2020gan}.

We include further related work on synthetic data generation and robustness datasets in \cref{appendix:related-work}.

\begin{table}[h]
    \centering
    \footnotesize{
     \begin{tabular}{p{0.125\linewidth} p{0.625\linewidth} p{0.1\linewidth}} \\ 
     \toprule
\textsc{\textsc{F.O.V.}}	 & \textsc{Dataset Configuration} & \textsc{Images} \\ \midrule
\textsc{Size} & Objects upright in the center, sizes from 1\% to 100\% of the image area in 1\% increments.& \num{92884}\\
\textsc{Location} & Objects upright. Sizes are 20\% of the image area. We do a grid search of locations, dividing the $x$-coordinate and $y$-coordinate dimensions into 20 equal parts each, for a total of 441 (21$\times$21) coordinate locations. & \num{479184}\\
\textsc{Rotation} & Objects in the center, sizes equal to 20\%, 50\%, 80\% or 100\% of the image size. Rotation angles ranging from 1 to 341 degrees counterclockwise in 20-degree increments.& \num{39540}\\
     \toprule
     \end{tabular}
     }
    \caption{Synthetic dataset details. The first column shows the relevant factor of variation (F.O.V.). When there are multiple values for multiple factors of variation, we generate the full cross product. 
    } \label{table:synth-data-description}
\end{table}

\section{Synthetic dataset details}
\vspace{2mm}
To construct our datasets, we paste foreground images (images of objects) on uncluttered background images in a precise way according to what we wish to study. The foregrounds are extracted from OpenImages~\citep{OpenImages} using the provided segmentation masks. We include only object classes that map to \imagenet classes. We also remove all objects that are tagged as occluded or truncated, and manually remove highly incomplete or inaccurately labeled objects. We manually filter the backgrounds to remove those with prominent objects, such as images focused on a single animal or person. This results in 614 object instances across 62 classes and 867 backgrounds.

We construct three subsets for evaluation, one corresponding to each factor of variation, as shown in \Cref{table:synth-data-description}. We provide further details in \cref{appendix:synthetic-dataset-details}.

\section{Results}

Using this dataset, we quantify fine-grained model robustness and uncover insights about models. Here we discuss three main groups of models we investigated: ResNets, Vision Transformers and CLIP. We investigate robustness to different object locations, sizes and rotation angles and include only highlights in the main paper. For full results, please see \cref{appendix:full-results}.

\subsection{ResNets}

\begin{figure*}
    \centering
    \begin{subfigure}[b]{0.4\textwidth}
        \centering
        \hspace{-0.1cm}\includegraphics[width=\linewidth]{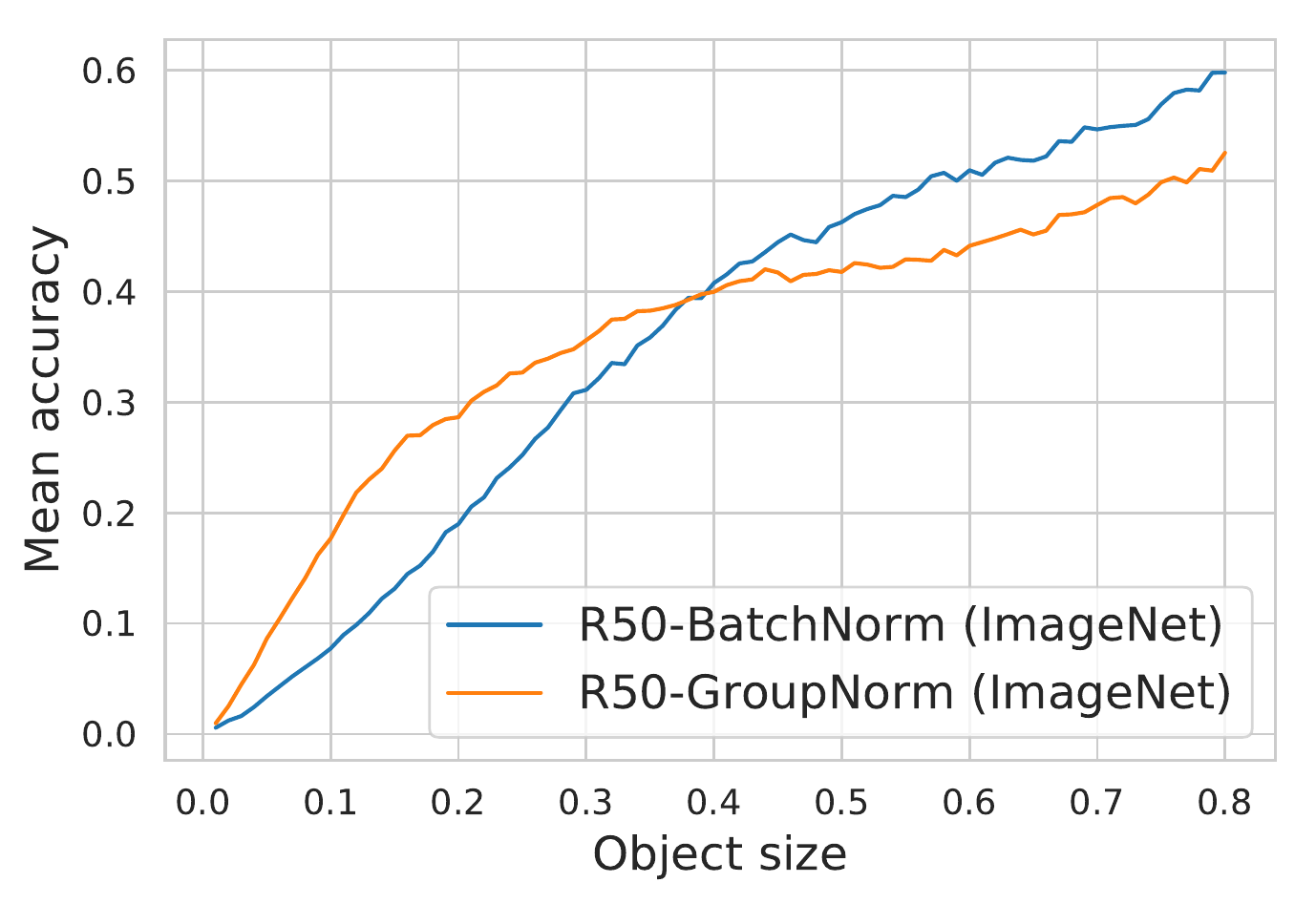}
    \end{subfigure}\hfill
    \begin{subfigure}[b]{0.54\textwidth}  
        \centering 
        \includegraphics[width=\linewidth]{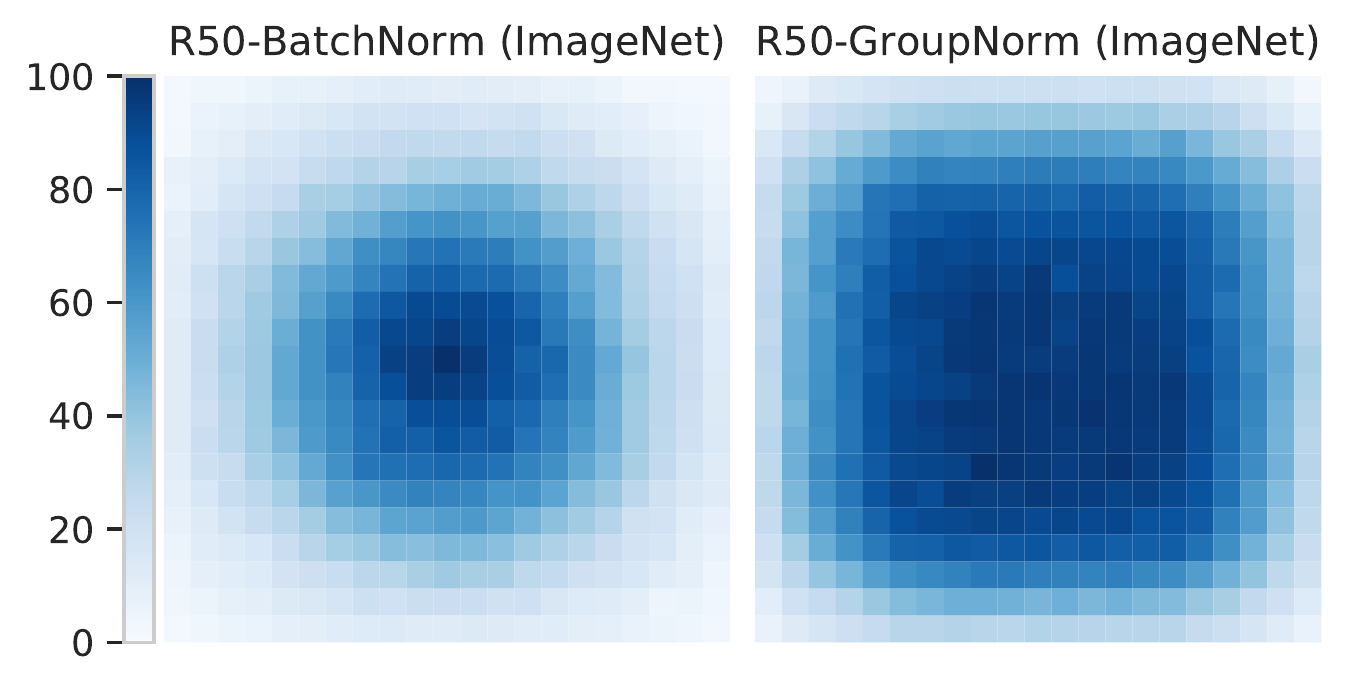}
    \end{subfigure}
    \begin{subfigure}[b]{0.6\textwidth}  
        \centering 
        \includegraphics[width=\linewidth]{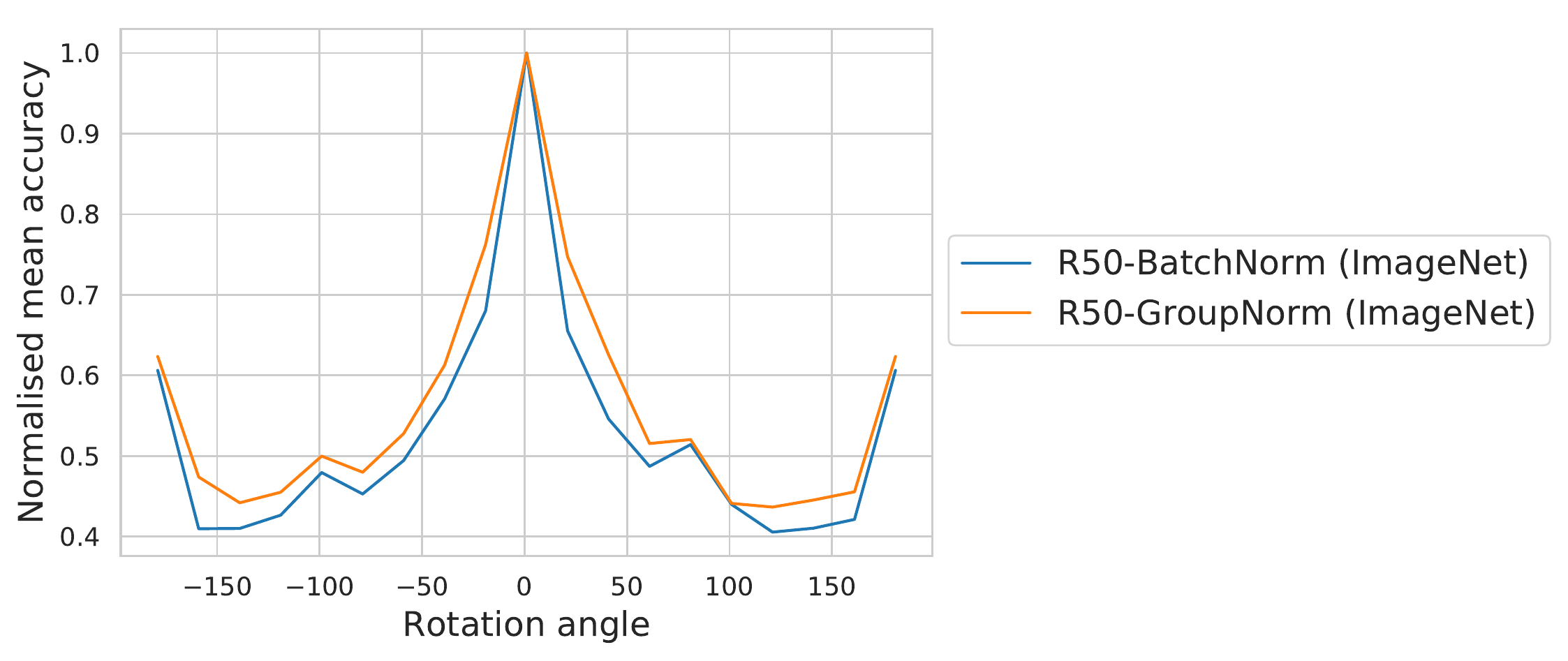}
    \end{subfigure}
    \vspace{1.5mm}
    \caption{\textbf{Left}: We find that a ResNet-50 using GroupNorm has higher accuracy on smaller objects, whereas the one using BatchNorm has higher accuracy on large objects. \textbf{Right}: The ResNet-50 using BatchNorm is less robust to changes in location than the one using GroupNorm. Each pixel represents the average normalised top-1 accuracy of the model on images where the object is centered at that location. The accuracy is shown as a percentage of the maximum accuracy across all locations. \textbf{Bottom}: The ResNet-50 using BatchNorm seems to be slightly less robust to changes in rotation angle than the one using GroupNorm.} \label{fig:r50_bngn} \vspace{-4mm}
\end{figure*}

ResNets \citep{he2016deep} are commonly-used architectures in computer vision. There are many decision choices involved, one of which is the normalisation method. The first widely adopted version used BatchNorm \citep{ioffe2015batch}, but GroupNorm \citep{wu2018group} has also been a popular choice since. We analyse ResNet-50 models pre-trained on \imagenet that use BatchNorm and GroupNorm respectively, and find three qualitative differences between them. First, the model that uses GroupNorm has higher accuracy on smaller objects, whereas the model that uses BatchNorm has higher accuracy on objects that take up at least 40\% of the image (\cref{fig:r50_bngn} left). Note that most objects usually take up less than 40\% of the image - for example, in a self-driving scenario, each object of interest in the driver's field of view typically occupies less than 40\% of it. Because of this, this tradeoff would generally be more favorable for the ResNet-50 that uses GroupNorm. Second, the model that uses BatchNorm is less robust to changes in location than the model using GroupNorm (\cref{fig:r50_bngn} right). For this experiment we use objects occupying 20\% of the image. Thirdly, it seems that ResNets using GroupNorm are also slightly more robust to changes in object orientation (rotation angle) (\cref{fig:r50_bngn} bottom). Note that we measure the robustness \emph{relative} to the model's best accuracy across locations or rotation angle respectively, so differences in absolute accuracy are accounted for. In future work, we hope to investigate whether these differences are present at scale when training on larger datasets and architectures.

\subsection{Vision Transformers}

Recently, \citet{dosovitskiy2020vit} showed that Transformers can be effective for image classification. Since the Vision Transformer (ViT) models use image patches as input, a natural question is whether they are robust to changes in object location, and whether they exhibit a grid-like pattern in per-location accuracy. We compare Vision Transformer models with convolutional neural networks with similar \imagenet accuracy. Specifically, we use BigTransfer (BiT) \citep{bit} models, which are ResNets using GroupNorm that were pre-trained on the same datasets and fine-tuned on \imagenet with high resolution. Absolute robustness is highly correlated with \imagenet accuracy \citep{taori2020measuring, djolonga2020robustness}, therefore, we compare models with similar \imagenet accuracy to account for this confounder.

\begin{figure}
\floatbox[{\capbeside\thisfloatsetup{capbesideposition={right,top},capbesidewidth=.5\textwidth}}]{figure}[\FBwidth]
{\caption{For each location on the grid, we compute the average accuracy on images with the object centered at that location. We show the accuracy as a percentage of the maximum accuracy across all locations. The third column indicates the difference between the second and first column. Blue indicates an improvement of the second column over the first column. This shows the difference in robustness to changes in object location between the pairs of models. Note that we compare models that have \emph{similar \imagenet accuracy}. \textbf{Rows 1 and 2}: We observe that the ViT models are more robust to location near the edges than the BiT ResNet models, as shown by the dark blue edges. We use a finer grid to investigate whether ViT models have grid-like patterns in location robustness, and do not find such patterns. \textbf{Row 3}: We observe that the CLIP model is slightly more robust to location despite having much lower \imagenet accuracy than the vanilla ResNet-50 model.}\label{fig:vit}}
{\begin{subfigure}[b]{.5\textwidth}
\includegraphics[width=\linewidth]{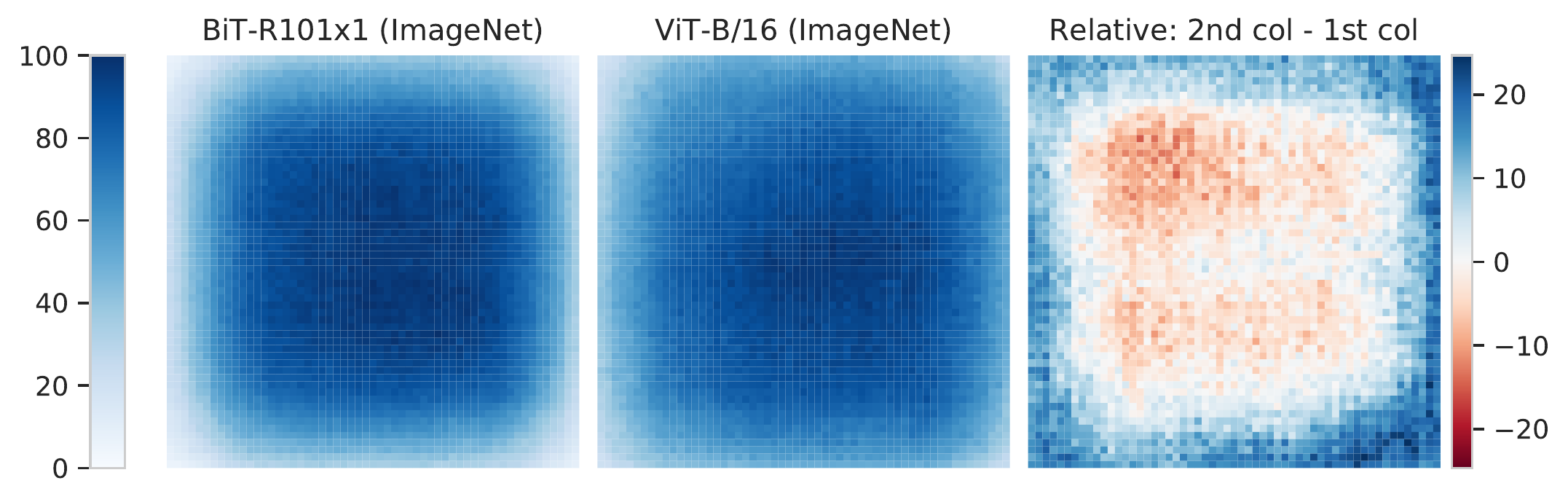} \\
\hspace{-0.1cm}\includegraphics[width=\linewidth]{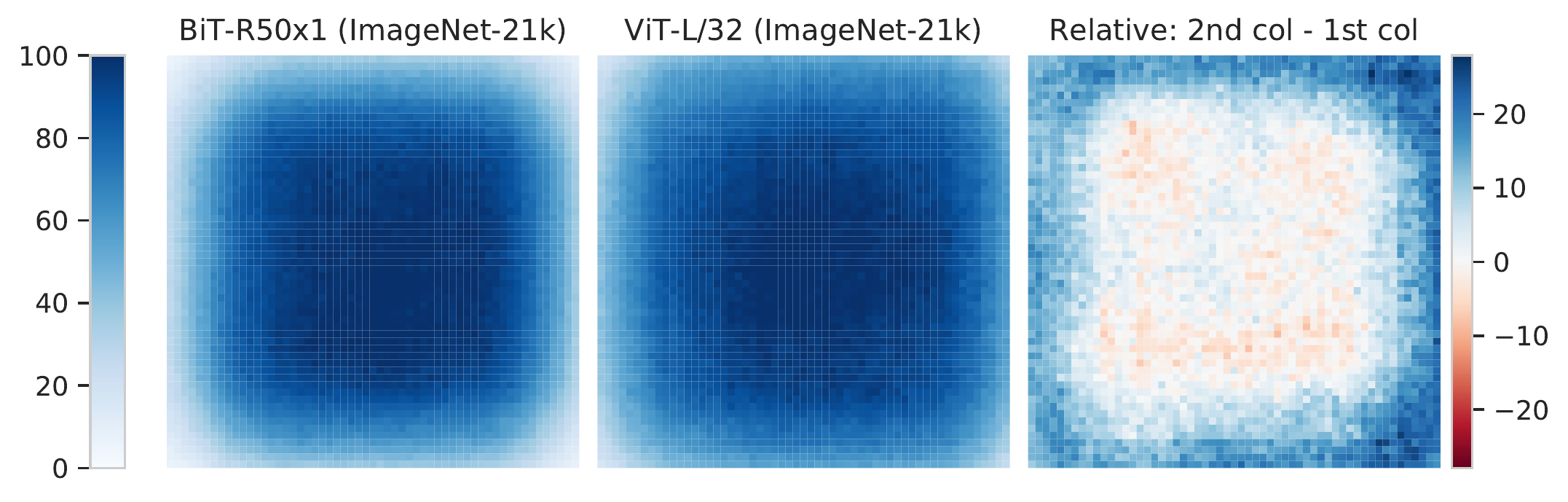} \\
\hspace{-0.1cm}\includegraphics[width=\linewidth]{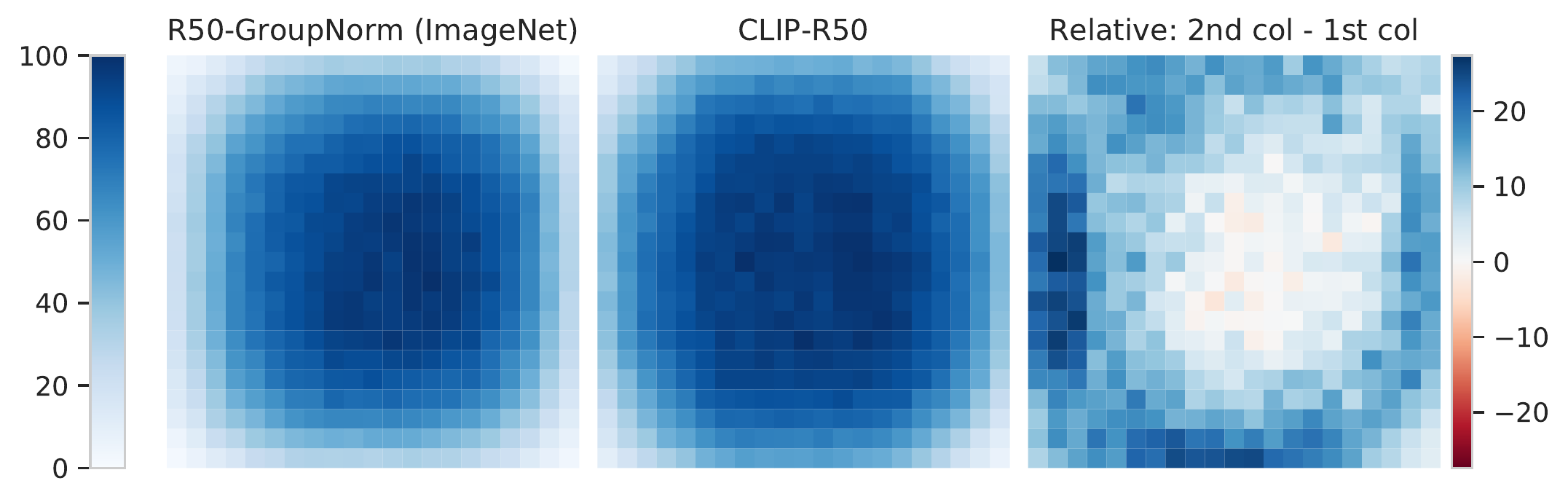} \\
\end{subfigure}}
\end{figure}

First, we find that ViT models have much higher relative accuracy when the object of interest is placed close to the edges of the image (\cref{fig:vit} rows 1 and 2). One potential explanation is that BiT models have zero padding in the convolutions, whereas ViT models do not have such padding. At the same time, in non-central, non-edge parts of the image, ViT models seem to be slightly more location-robust than BiT models in most but not all locations.

Second, we do not find evidence for a grid-like pattern in per-location accuracy in ViT. To investigate this, we use a finer $56\times 56$ grid compared to the $20\times 20$ grid in the previously open-sourced dataset. The ViT-*/16 models use image patches forming a $14\times 14$ grid, so each patch would correspond to a $4\times 4$ patch on this grid (\cref{fig:vit} row 1).  The ViT-*/32 models use a $7\times 7$ grid, so each patch would correspond to a $8 \times 8$ patch on this grid (\cref{fig:vit} row 2). When looking at the absolute values and differences in accuracies, we do not see grid-like patterns. Thus, it seems that there are no significant grid-like patterns in per-location accuracy in ViT.

\subsection{CLIP}

One model that has received a great deal of attention is CLIP \citep{2021clip}. It stands in contrast to other considered models since it was trained jointly on images and language input. Furthermore, in contrast to other models, CLIP is not fine-tuned to \imagenet, but is evaluated in the \imagenet label space in a zero-shot setting. As a result, its \imagenet accuracy --- at least of the small published models and without prompt ensembling --- is significantly lower than that of the other models in this paper. Notably, even when we compare CLIP models to a standard ResNet-50 that has over 10\% higher top-1 accuracy on \imagenet, the CLIP model seems to be more robust to different object locations (\cref{fig:vit} row 3). This is perhaps surprising since robustness is often correlated with ImageNet accuracy, but is in line with its improved relative performance on robustness benchmarks, as reported in~\citet{2021clip}.

\section{Discussion and Conclusion}
We investigated robustness of ResNets, Vision Transformers and CLIP to changes in object location and size. Additional results on robustness with respect to object rotation can be found in \cref{appendix:full-results}.

We note that there could be potential differences and confounding factors when evaluating the performance on synthetic data. We apply the following steps to mitigate the risk: Firstly, we use cut-and-pastes of real data instead of fully synthetic data. Notably, \citet{ghiasi2020copypaste} successfully trained state-of-the-art object segmentation models on such data, which lends evidence that related artifacts may not significantly affect behaviour. Secondly, we average across over 1000 object and background combinations to minimize the effect of the choice of object or background on the results. Finally, we consider relative performance between models as opposed to absolute numbers.

We hope that the insights presented in this study will influence  research on the use of synthetic data for stress-testing deep learning models.

\bibliography{iclr2021_conference}
\bibliographystyle{iclr2021_conference}

\newpage
\appendix

\section{Further Related Work}
\label{appendix:related-work}
In this section, we describe related work on synthetic data generation and datasets to measure robustness.

Other efforts on synthetic data generation include CLEVR \citep{DBLP:conf/cvpr/JohnsonHMFZG17}, which aims to evaluate compositional generalisation, dSprites~\citep{dsprites17}, which aims to evaluate disentanglement of latent features, or smallNorb~\citep{DBLP:conf/cvpr/LeCunHB04}, which is also an object classification dataset, albeit with different factors of variation except rotation. These datasets use rendered shapes or models of geometric shapes or toys instead of realistic photos of ImageNet classes with photo backgrounds.

Our work focuses on synthetic data to analyse specific factors of variation. Other datasets to analyse robustness mostly include natural datasets. For example, ImageNet-R presents a dataset of alternatively rendered imagery ranging from cartoons to origami ~\citep{hendrycks2020many}. ImageNet-Vid~\citep{ILSVRC15} uses frames from video sequences, and ImageNet-Vid-Robust measures whether model predictions are correct and consistent across similar frames~\citep{imnetvid}. Finally, ImageNet-C~\citep{hendrycks2018benchmarking} uses synthetic image-level perturbations on natural images to analyse robustness with respect to perturbations such as Gaussian noise, JPEG compression, variations in image brightness or motion blur. \textsc{SI-Score} focuses on object-level as opposed to image-level factors of variation.

\section{Synthetic dataset details}
\label{appendix:synthetic-dataset-details}

We include further details on the synthetic dataset in this section.

We construct three subsets for evaluation, one corresponding to each factor of variation we wanted to investigate (object size, location and rotation), as shown in \Cref{table:synth-data-description2}. The table is repeated here for easy reference. For each object instance, we sample two backgrounds, and for each of these object-background combinations, we take a cross product over all the factors of variation. For the datasets with multiple values for more than one factor of variation, we take a cross product of all the values for each factor of variation in the set. For example, for the rotation angle dataset, there are four object sizes and 18 rotation angles, so we do a cross product and have 72 factor of variation combinations. For the object size and rotation datasets, we only consider images where objects are at least 95\% in the image. For the location dataset, such filtering removes almost all images where objects are near the edges of the image, so we do not do such filtering. Note that since we use the central coordinates of objects as their location, at least 25\% of each object is in the image even if we do not do any filtering.

\begin{table}[h]
    \centering
    \footnotesize{
     \begin{tabular}{p{0.125\linewidth} p{0.625\linewidth} p{0.1\linewidth}} \\ 
     \toprule
\textsc{\textsc{F.O.V.}}	 & \textsc{Dataset Configuration} & \textsc{Images} \\ \midrule
\textsc{Size} & Objects upright in the center, sizes from 1\% to 100\% of the image area in 1\% increments.& \num{92884}\\
\textsc{Location} & Objects upright. Sizes are 20\% of the image area. We do a grid search of locations, dividing the x-coordinate dimension and y-coordinate dimensions into 20 equal parts each, for a total of 441 (21$\times$ 21) coordinate locations. & \num{479184}\\
\textsc{Rotation} & Objects in the center, sizes equal to 20\%, 50\%, 80\% or 100\% of the image size. Rotation angles ranging from 1 to 341 degrees counterclockwise in 20-degree increments.& \num{39540}\\
     \toprule
     \end{tabular}
     }
    \caption{Synthetic dataset details. The first column shows the relevant factor of variation (F.O.V.). When there are multiple values for multiple factors of variation, we generate the full cross product of images.} \label{table:synth-data-description2}
\end{table}

\textbf{Image licenses} \quad The backgrounds are images from nature taken from \emph{pexels.com}. The license therein allows one to reuse photos with modifications.

\newpage
\section{Full results for model comparisons in the main paper}
\label{appendix:full-results}

\subsection{ResNet-50s with BatchNorm and GroupNorm}
\label{appendix:r50}

\begin{figure*}[h]
    \centering
    \begin{subfigure}[b]{\textwidth}
        \centering
        \includegraphics[width=0.7\linewidth]{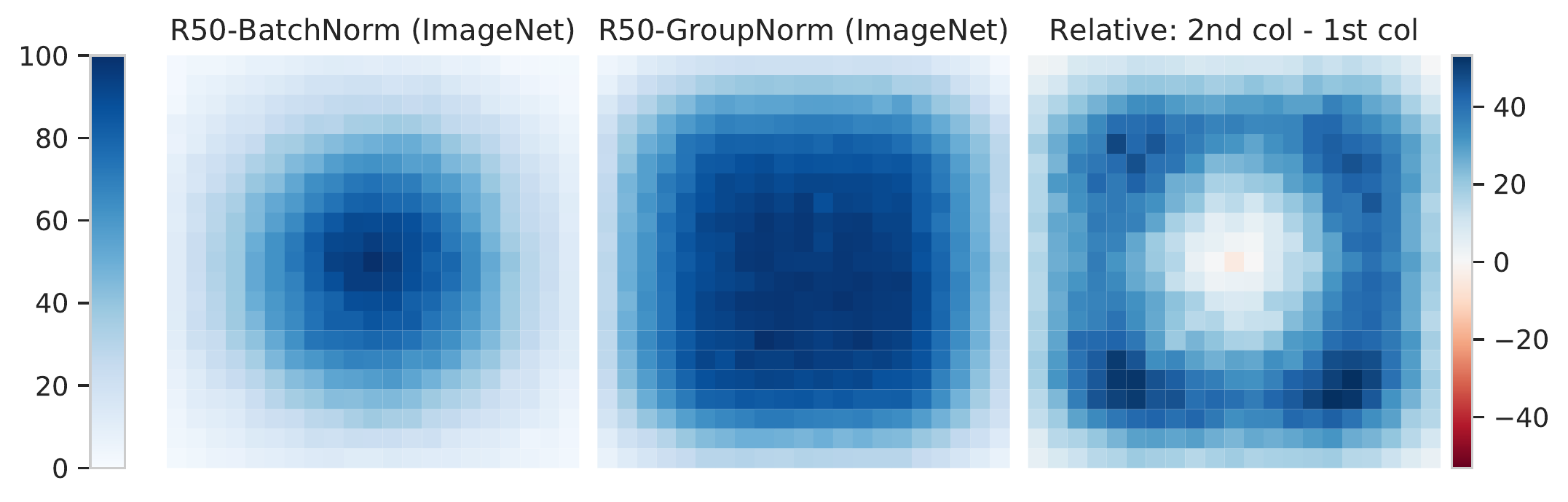} \\
    \end{subfigure}\hfill
    \vspace{1.5mm}
    \caption{In the first and second columns, for each location on the grid, we compute the average accuracy of the models on images where the object is centered at that location. We show the accuracy as a percentage of the maximum accuracy across all locations. In the third column, we compute the difference between the second column and the first column. Blue indicates an improvement of the second column over the first column. This shows the difference in robustness to changes in object location between the two models. The ResNet-50 that uses BatchNorm is less robust to changes in location than the one that uses GroupNorm. } \vspace{-4mm}
\end{figure*}

The object size and rotation plots are in Figure~\ref{fig:r50_bngn} in the main paper.

\newpage
\subsection{BiT (ResNet) vs ViT models}
\label{appendix:bit-vit}

We compare BigTransfer (BiT) \citep{bit} and Vision Transformer (ViT) \citep{dosovitskiy2020vit} model pairs that have similar \imagenet accuracy. We include a ViT-L/32 model in the location study to investigate whether there is a grid-like pattern in location robustness.

\begin{figure}[h]
    \centering
    \begin{subfigure}[b]{\textwidth}
        \centering
        \includegraphics[width=0.7\linewidth]{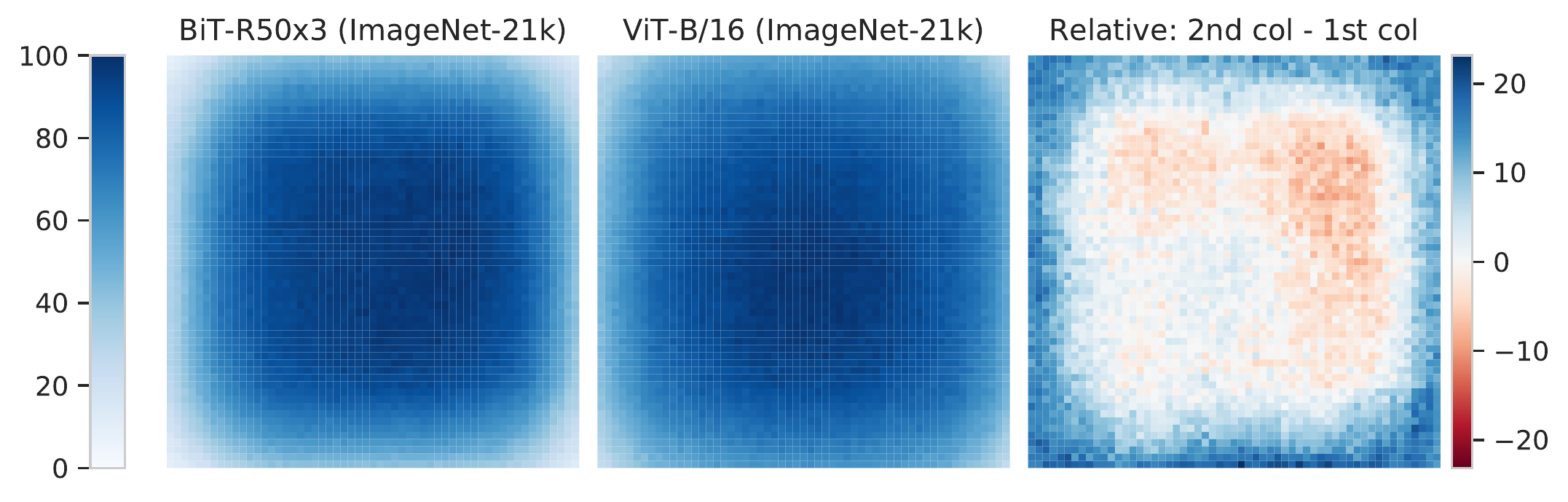} \\
    \end{subfigure}\hfill
    \vspace{1.5mm}
    \caption{In the first and second columns, for each location on the grid, we compute the average accuracy of the models on images where the object is centered at that location. We show the accuracy as a percentage of the maximum accuracy across all locations. In the third column, we compute the difference between the second column and the first column. Blue indicates an improvement of the second column over the first column. This shows the difference in robustness to changes in object location between the pairs of models. We observe that the ViT models are more robust to location near the edges than the BiT ResNet models, as shown by the dark blue edges in the third column.  We use a finer grid to investigate whether ViT models have grid-like patterns in location robustness, and do not find such patterns. Note that we compare models that have similar \imagenet accuracy.} \vspace{-4mm}
\end{figure}

The plots comparing the location robustness of BiT-R101x3 with ViT-B/16 trained on ImageNet, and BiT-R50x1 with ViT-L/32 trained on ImageNet-21k are in Figure~\ref{fig:vit} (top, middle rows) in the main paper. We compare these pairs of models that have similar ImageNet accuracy to control for differences purely due to different ImageNet accuracy.

\begin{figure}[h]
    \centering
    \begin{subfigure}[b]{0.38\textwidth}
    \includegraphics[width=\textwidth]{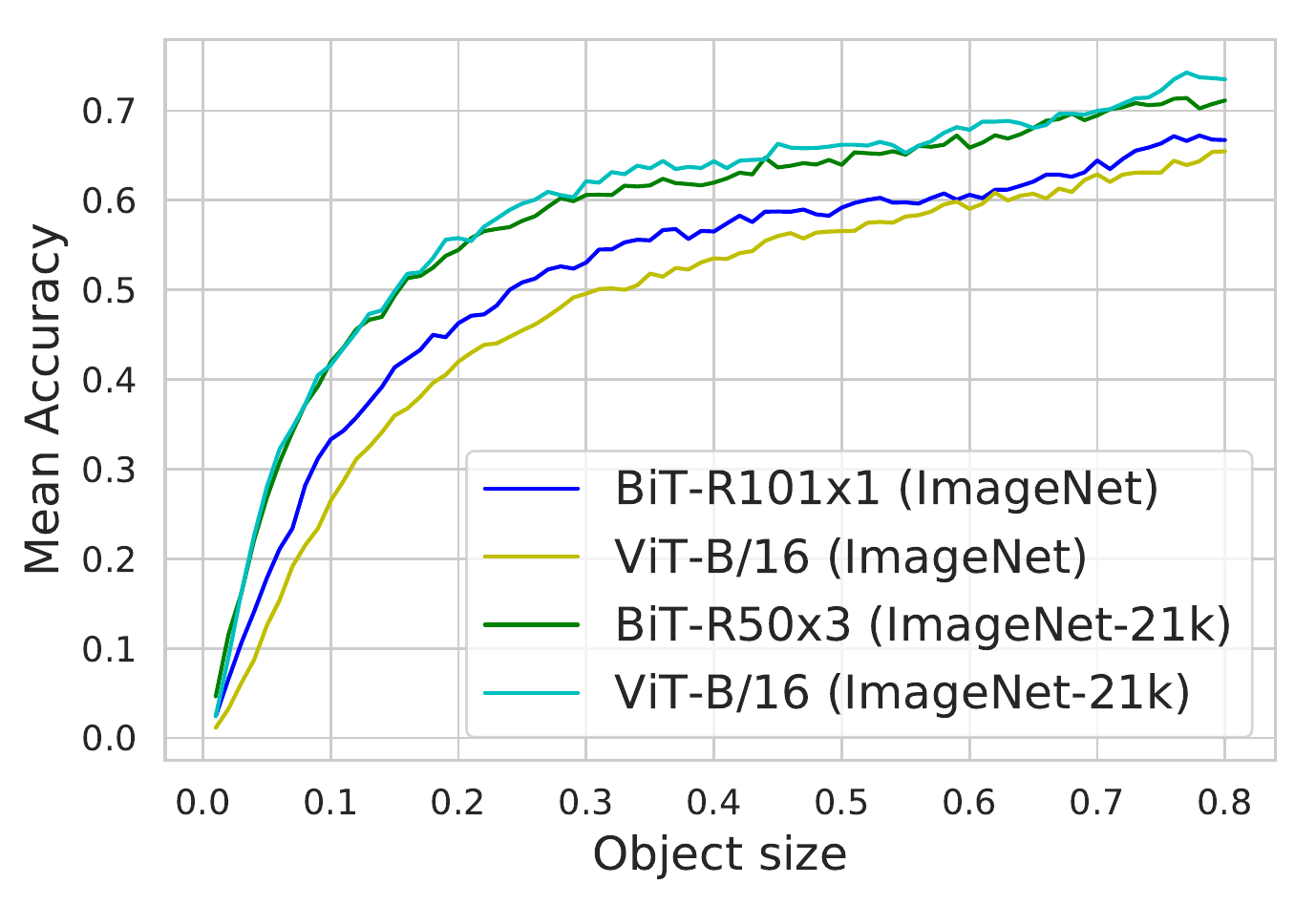}
    \end{subfigure}\hfill
    \begin{subfigure}[b]{0.6\textwidth}
    \includegraphics[width=\textwidth]{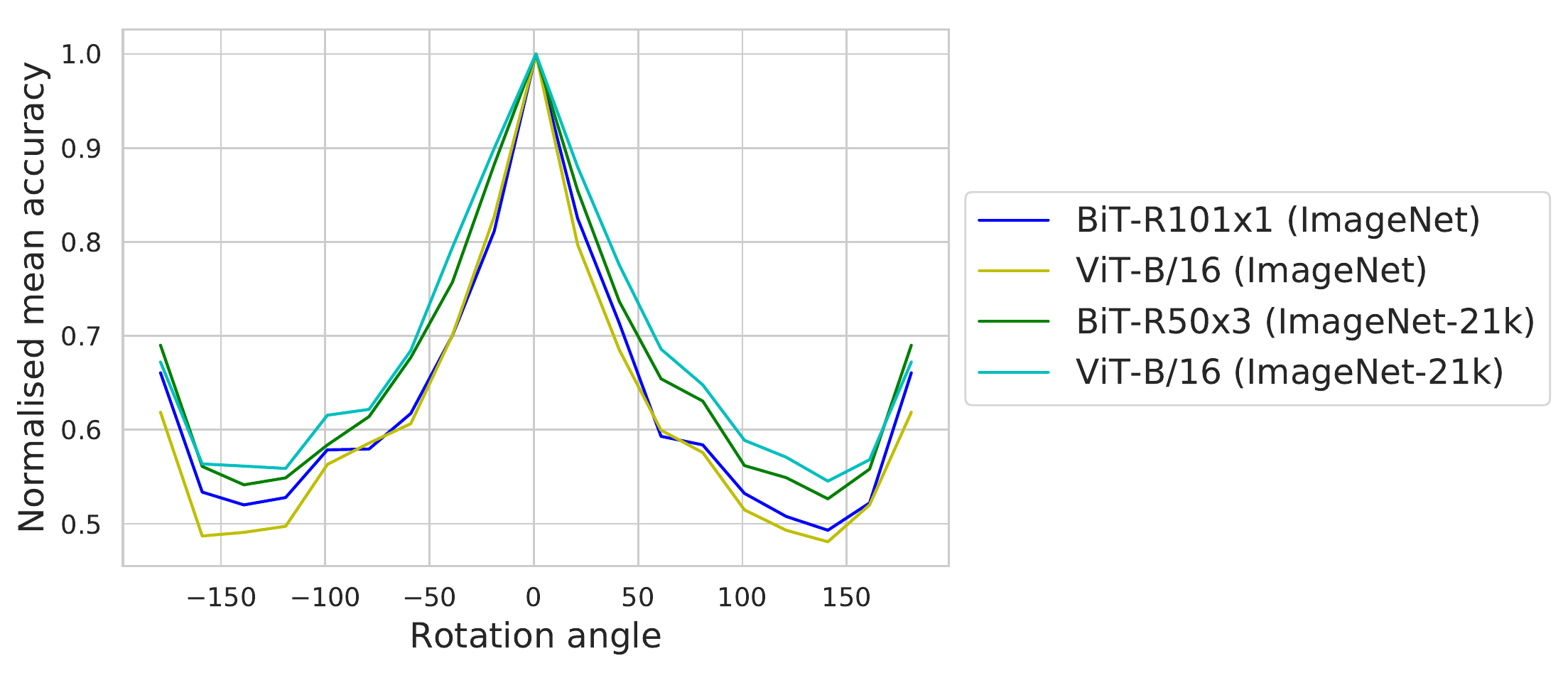}
    \end{subfigure}
    \caption{ \textbf{Left}: We find that the BiT-R50x3 and ViT-B/16 models trained on ImageNet-21k seem to have similar robustness to changes in object size. However, the BiT-R101x1 model trained on ImageNet seems to be slightly better at classifying smaller objects than the ViT-B/16 model trained on ImageNet. \textbf{Right}: Conversely, the two BiT and ViT models trained on ImageNet seem to have similar robustness to changes in rotation angles with the BiT model perhaps being slightly better. For the two models trained on ImageNet-21k, however, the ViT model seems to be slightly more robust. The differences in both cases are quite small.}
    \label{}
\end{figure}

\newpage
\subsection{CLIP models \citep{2021clip}}
\label{appendix:clip}

\begin{figure*}[h]
    \centering
    \begin{subfigure}[b]{\textwidth}
        \centering
        \includegraphics[width=0.7\linewidth]{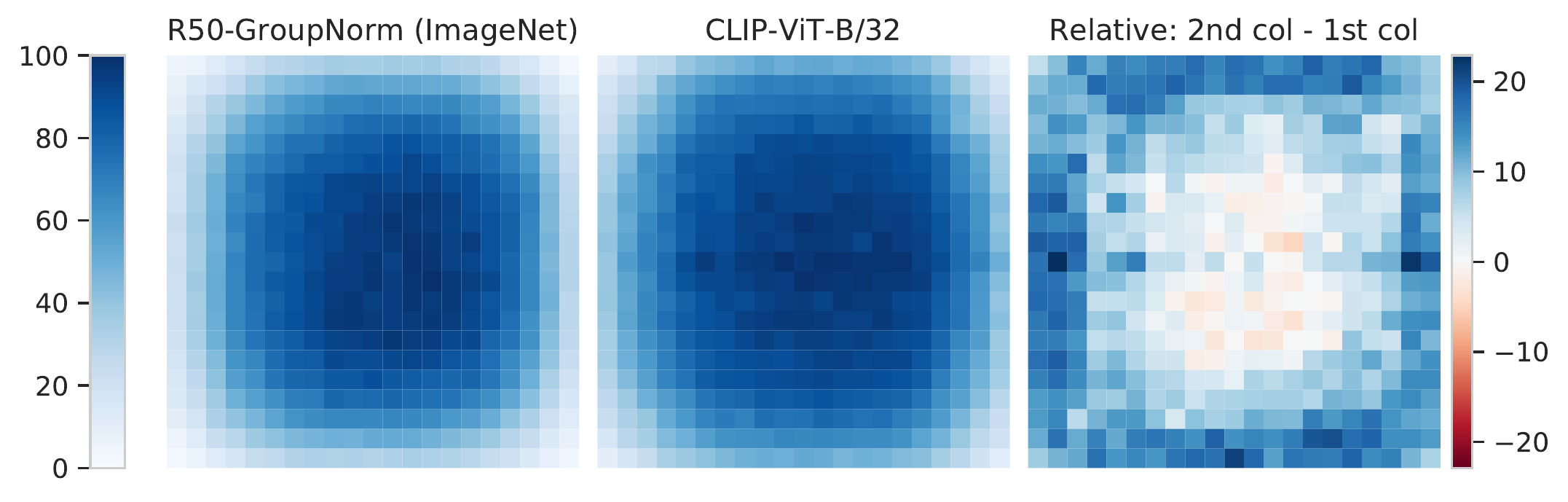} \\
    \end{subfigure}\hfill
    \vspace{1.5mm}
    \caption{In the first and second columns, for each location on the grid, we compute the average accuracy of the models on images where the object is centered at that location. We show the accuracy as a percentage of the maximum accuracy across all locations. In the third column, we compute the difference between the second column and the first column. Blue indicates an improvement of the second column over the first column. This shows the difference in robustness to changes in object location between the pairs of models. We observe that the CLIP model is slightly more robust to location despite having much lower \imagenet accuracy than the vanilla ResNet-50 model, as can be seen by the blue edges in the third column.} \vspace{-4mm}
\end{figure*}

The plot comparing the location robustness of the ResNet-50 (GroupNorm) with the CLIP model using a ResNet-50 backbone is in Figure~\ref{fig:vit} (bottom row) in the main paper.

\begin{figure}[h]
    \centering
    \begin{subfigure}[b]{0.4\textwidth}
    \includegraphics[width=\textwidth]{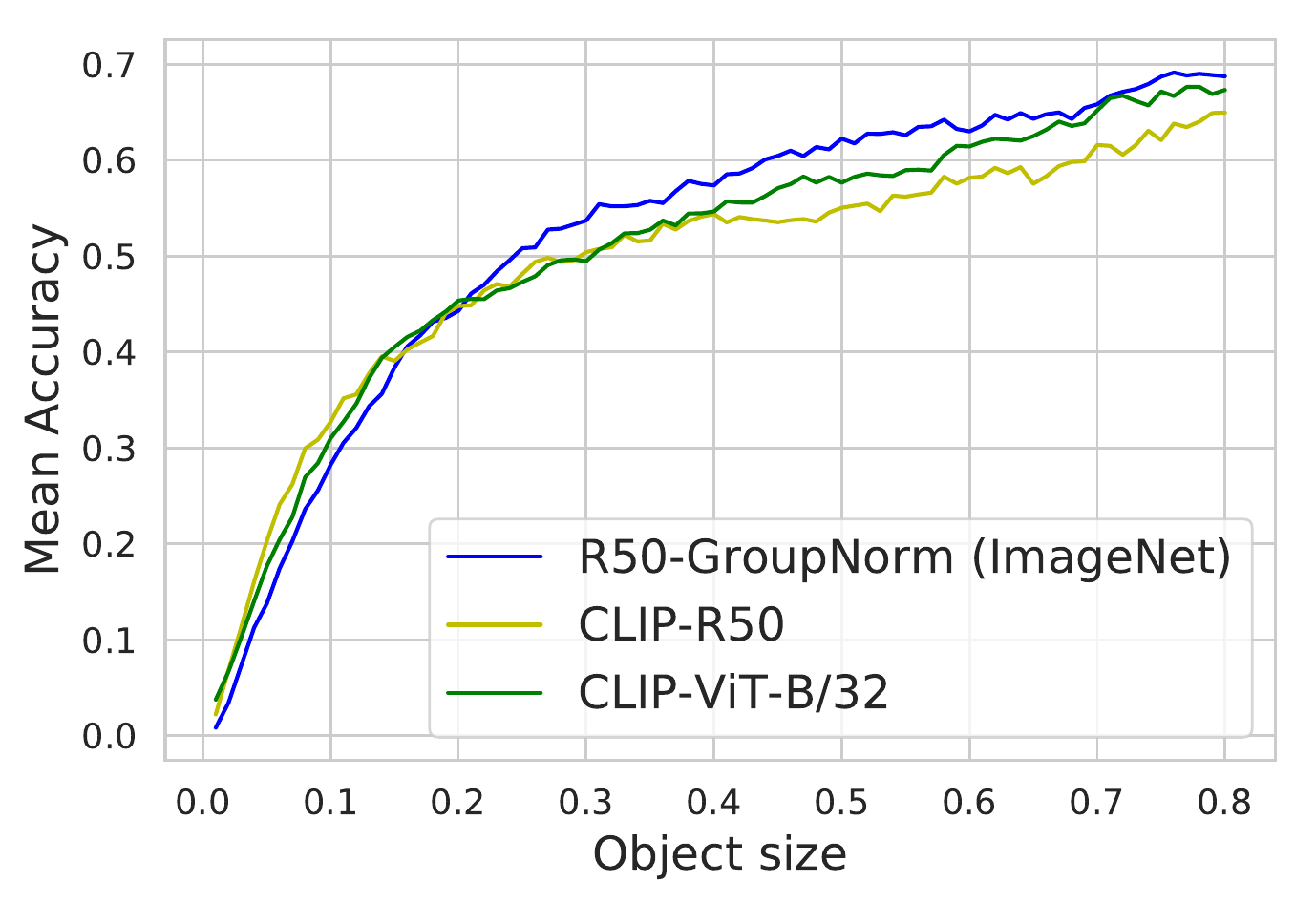}
    \end{subfigure}
    \begin{subfigure}[b]{0.4\textwidth}
    \includegraphics[width=\textwidth]{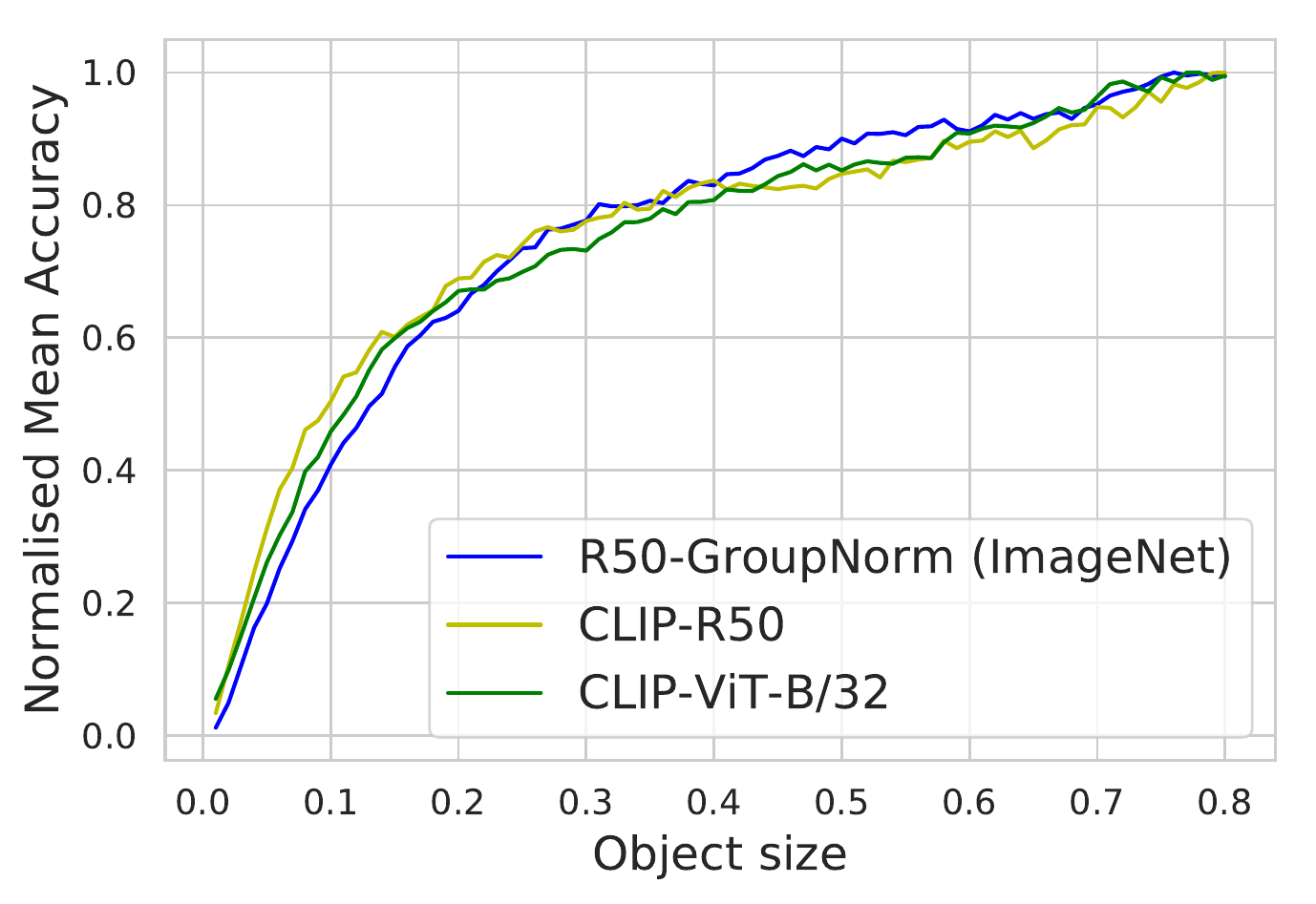}
    \end{subfigure}\hfill
    \caption{Note that the CLIP models have lower ImageNet accuracy than the ResNet-50 model. We used the ResNet-50 model because were not able to find standard models with lower ImageNet accuracy. Given the large difference in ImageNet accuracy between these three models, we plot the normalised accuracy (accuracy as a percentage of the highest accuracy per model across all sizes) as well (right). The plots suggest that the CLIP models may be slightly more robust than the vanilla ResNet-50 on small objects, and the vanilla ResNet-50 may be more robust on medium-sized objects.}
    \label{}
\end{figure}

\begin{figure}[h]
    \centering
    \begin{subfigure}[b]{0.6\textwidth}
    \includegraphics[width=\textwidth]{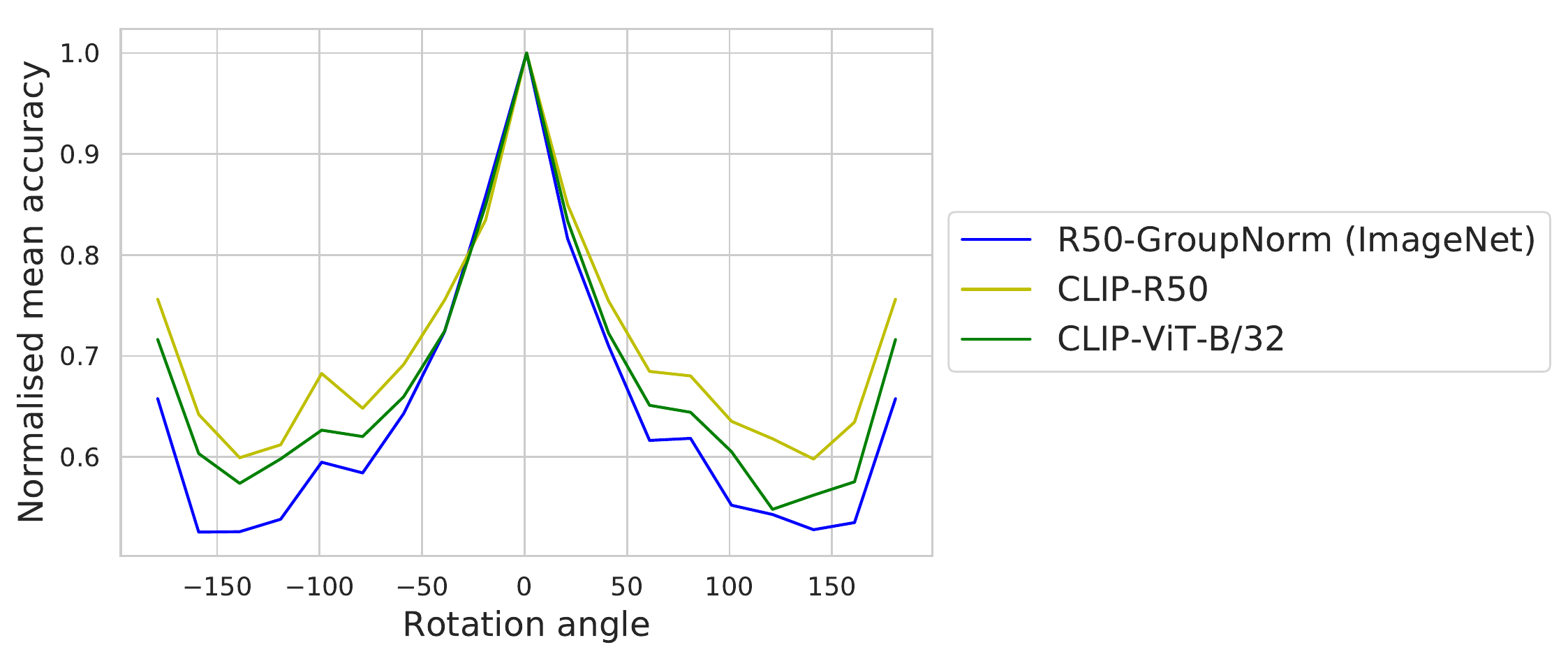}
    \end{subfigure}
    \caption{The CLIP models are more robust to changes in object rotation angle than the ResNet 50, with the CLIP-R50 model being more robust than the CLIP-ViT-B/32 model.}
    \label{}
\end{figure}

\end{document}